\documentclass[letterpaper, 10 pt, conference]{ieeeconf} 

\IEEEoverridecommandlockouts
\usepackage{tabularx, graphicx,
  amsmath,amsfonts,siunitx,pifont,amssymb}
\usepackage[table]{xcolor}
\usepackage{xcolor, bm, float, multirow, multicol, courier, mathtools, url}
\usepackage[ruled,vlined]{algorithm2e} \pdfminorversion=4
\let\labelindent\relax \usepackage{enumitem} \usepackage{graphicx}
\usepackage{adjustbox}
\usepackage{subcaption}
\usepackage{todonotes}

\newcommand{\etal}{\textit{et al}.~}

\newcommand{\abs}[1]{\lvert#1\rvert}

\usepackage{pdfpages}
\usepackage{cite}
\usepackage{balance}
\usepackage{hyperref}

\def\secref#1{Sec.~\ref{#1}}
\def\figref#1{Fig.~\ref{#1}}
\def\tabref#1{Table~\ref{#1}}
\def\eqref#1{Eq.~(\ref{#1})}
\def\algref#1{Alg.~\ref{#1}}
\def\appref#1{Appendix~\ref{#1}}

\title{\LARGE \bf{Scalable and Elastic LiDAR Reconstruction in\\Complex Environments Through Spatial
Analysis}}

\author{Yiduo Wang\textsuperscript{1}, Milad Ramezani\textsuperscript{1}, Matias Mattamala\textsuperscript{1} and Maurice Fallon\textsuperscript{1}
    \thanks{This research is supported by the ESPRC ORCA Robotics Hub (EP/R026173/1). M. Fallon is supported by a Royal Society University Research Fellowship.}
    \thanks{\textsuperscript{1} These authors are with the Oxford Robotics Institute, University of Oxford, UK.
        {\tt\small \{ywang, milad, matias, mfallon\}@robots.ox.ac.uk}
    \newline 978-1-6654-1213-1/21/\$31.00 \textcopyright 2021 IEEE}%
}


\begin{document}
	
\setlength{\abovedisplayskip}{4pt}
\setlength{\belowdisplayskip}{4pt}
	
\maketitle 
\thispagestyle{empty} 
\pagestyle{empty}
	

\begin{abstract}
This paper presents novel strategies for spawning and fusing submaps 
within an elastic dense 3D reconstruction system. 
The proposed system uses spatial understanding of the scanned environment 
to control memory usage growth by fusing overlapping submaps in different ways.
This allows the number of submaps and memory consumption 
to scale with the size of the environment rather than the duration of exploration.
By analysing spatial overlap, our
system segments distinct spaces, 
such as rooms and stairwells on the fly during exploration.
Additionally, we present a new mathematical formulation of relative uncertainty
between poses to improve the global consistency of the reconstruction. 
Performance is demonstrated using a multi-floor multi-room indoor experiment, 
a large-scale outdoor experiment and simulated datasets. 
Relative to our baseline, 
the presented approach demonstrates improved scalability and accuracy.
\end{abstract}
	
\section{Introduction}

3D reconstruction is a common part of
applications such as active mapping~\cite{Bircher2018, Dai_ICRA2020}, 
collision avoidance~\cite{oleynikova2017voxblox} 
and the inspection of industrial assets~\cite{Hollinger2013, Franz2016}. 
Building Information Models (BIM) 
are commonly available for modern buildings, 
but there are scenarios where these models no longer 
represent the real situation, e.g. after renovations or disasters. 
Although systems have been developed to reconstruct 
these models offline using point clouds from
laser scanners, 
autonomous exploration and reconstruction 
in multi-storey environments is still an open challenge in mobile robotics.
This has been the motivation of
international competitions such as the
DARPA SubT challenge~\cite{bouman2020autonomous, ebadi2020lamp}.

\begin{figure}[!t]
    \centering
    \begin{subfigure}{\linewidth}
        \centering
        \includegraphics[width=0.9\linewidth,trim={0cm 0.5cm 0cm
            0.2cm},clip]{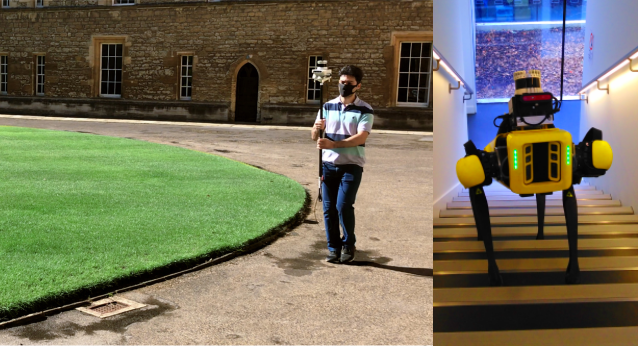}
    \end{subfigure}
    \begin{subfigure}{\linewidth}
        \centering
        \includegraphics[width=0.9\linewidth,trim={0cm 0cm 0cm
            -0.2cm},clip]{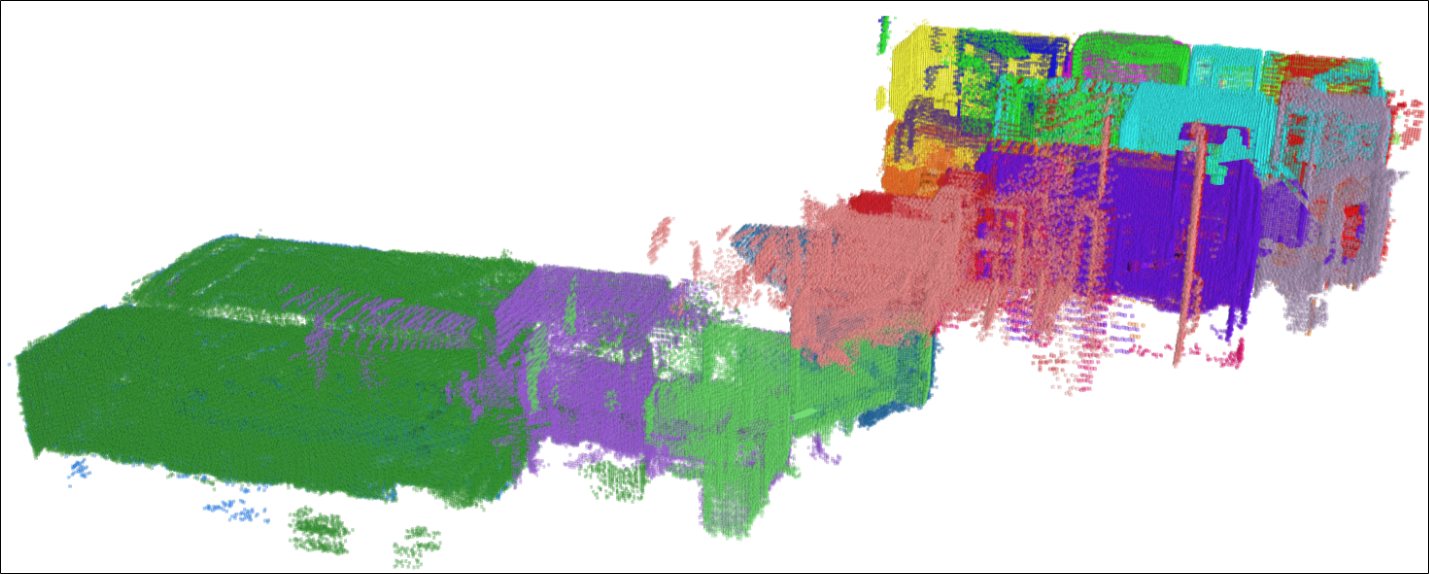}
    \end{subfigure}
    \begin{subfigure}{\linewidth}
        \centering
        \includegraphics[width=0.9\linewidth,trim={0cm 0cm 0cm
            -0.2cm},clip]{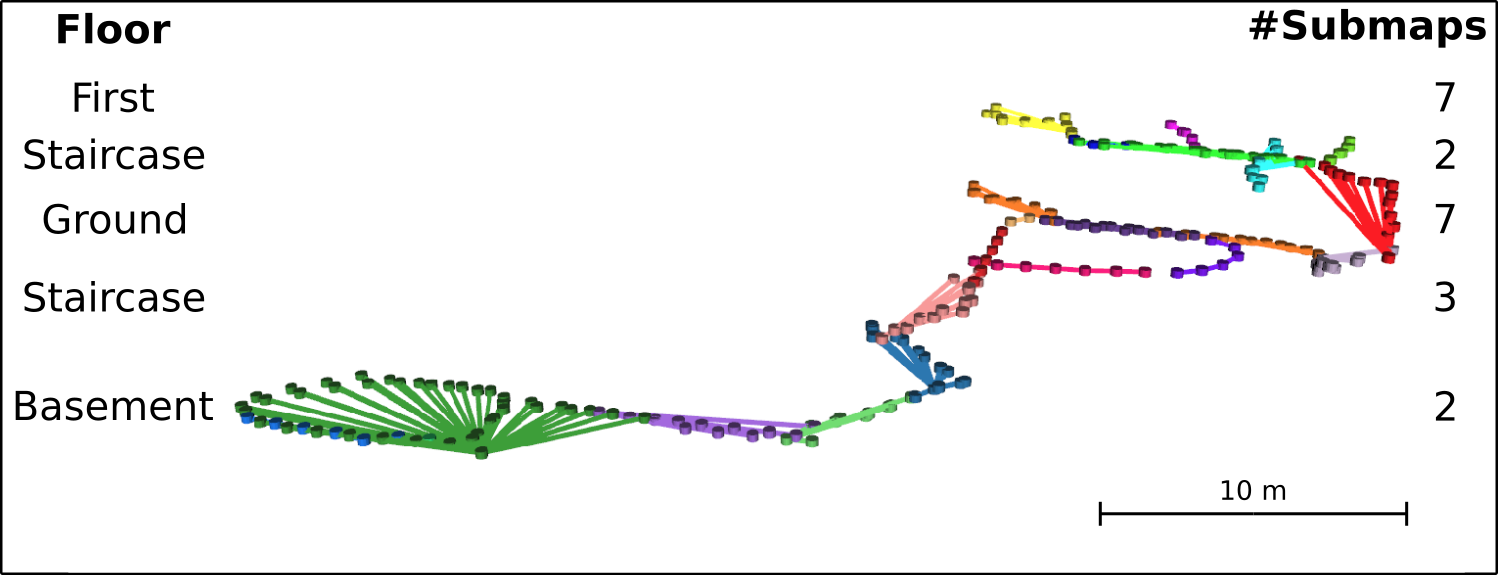}
    \end{subfigure}
    \caption{\small{Our proposed system has been tested in a large-scale outdoor and a multi-floor multi-room indoor environments. 
            \textbf{Top}: A handheld LiDAR system exploring New College, Oxford, UK (left) and the quadruped robot Spot climbing stairs in Oxford Robotics Institute (ORI) (right); 
            \textbf{Middle}: The reconstruction result of ORI, in which 
            each room is segmented by a submap indicated by a unique colour; 
            \textbf{Bottom}: The clustered pose graph nodes using spatial overlap analysis. 
            More demonstrations are available in the supplementary video: \url{https://youtu.be/QOc52EO3ULc}
    }}
    \label{fig:front_page_full_map}
        \vspace{-5mm}
\end{figure}

In this paper, we propose a dense 3D reconstruction system
which uses inputs from a 3D LiDAR on a mobile robot to map
complex environments, 
such as the three-storey building 
demonstrated in~\figref{fig:front_page_full_map}. 
Our system creates and maintains local occupancy submaps instead of a global map 
to account for loop closure corrections and
improve the global consistency of the reconstruction. 
We design strategies for 
spawning and fusing submaps based on 
geometric understanding of known spaces, 
enabling on-the-fly segmentation of 
areas that are isolated from one another, e.g. individual rooms indoors. 

The proposed system also leverages explicitly known free space
in its occupancy representation to analyse spatial overlap among submaps, 
and to fuse submaps together to reduce redundant reconstruction. 
In addition, by computing the relative uncertainty 
among poses in a SLAM system, 
our system rejects unreliable 
submap fusion and improves reconstruction accuracy. 
We demonstrate the performance of the proposed system using 
both simulation and real-world experiments, 
especially its improvement in reconstruction accuracy, 
system scalability and the capability of room segmentation. 

The contributions of our research are the following:
\begin{itemize}
    \item New strategies for spawning and fusing submaps using probabilistic and
spatial understanding.
    \item A new formulation for relative uncertainty derived from the work of
Mangelson~\etal\cite{mangelson2020characterizing} and GTSAM~\cite{dellaert2017gtsam}, 
and a formal treatment of uncertainty in submap 
fusion.
    \item Improved reconstruction accuracy and scalability in both outdoor
exploration experiments and a multi-storey/multi-room exploration with a
legged robot.
\end{itemize}

The overarching goal of the work is to achieve scalability with the size of the environment instead
of the
exploration length by controlling the growth of the submap number and
memory
consumption\footnote{Sparsification of the underlying SLAM pose graph is related research topic
which we do not explore in this work.}.

\begin{figure*}[t]
    \centering
    \includegraphics[width=0.9\linewidth]{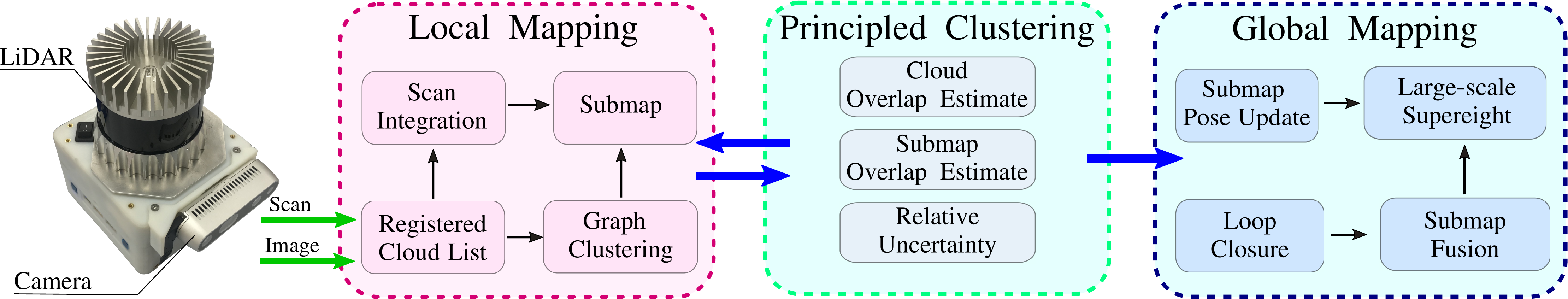}
    \caption{\small{An overview of the proposed system. 
            \textit{Principled Clustering} contains the new strategies based on cloud and submap overlap estimation as well as relative pose uncertainty in the SLAM graph. 
            This module interacts with \textit{Local Mapping} to spawn new submaps (\secref{sec:Spawning}), 
            and proposes or rejects submap fusion for \textit{Global Mapping} (\secref{sec:Fusion} and \secref{sec:Uncertainty}). }}
    \label{fig:system_overview}
        \vspace{-4mm}
\end{figure*}

\section{Related Work}
\label{sec:RelatedWork}
While the literature on dense 3D reconstruction is substantial, in this paper
we focus on systems incorporating submaps
and methods designed for room segmentation
because these aspects are the most relevant to our work. 

\subsection{3D Reconstruction via Submaps}
Submapping is a common technique in SLAM systems such as 
the \textit{Atlas} framework by Bosse~\etal\cite{bosse2003atlas} and 
DenseSLAM by Nieto~\etal\cite{Nieto2006DenseSLAM}. 
These SLAM systems reuse existing submaps for localisation 
when revisiting a known area rather than remapping that location. 
Maintaining a collection of submaps keeps memory and computation bounded 
as opposed to creating a single global map. 

Submaps can also enable
elasticity when a reconstruction requires correction at the event of a loop closure. 
Leveraging OctoMap-based~\cite{hornung2013octomap} submaps, 
Ho~\etal\cite{ho2018virtual} and Sodhi~\etal\cite{Sodhi2019} designed 
elastic occupancy maps. 
Voxgraph~\cite{reijgwart2020voxgraph} uses
the same strategy to create a globally consistent reconstruction
instead represented by a series of Signed Distance Function (SDF) submaps. 
In our previous work~\cite{wang2020elastic}, we developed
submap-based elastic reconstruction pipelines which supported 
both occupancy and SDF representations.

These approaches spawn new submaps after either a temporal interval~\cite{ho2018virtual, Sodhi2019, reijgwart2020voxgraph}
or distance travelled~\cite{wang2020elastic}
to bound the local odometry drift within each submap.
Extending upon~\cite{wang2020elastic}, this paper proposes submap spawning and fusion based on a
deeper spatial analysis to uniquely represent confined areas such as rooms.
This allows local planning to use only a submap in challenging exploration tasks such
as~\cite{Bellicoso2018AdvancesIR}.

\subsection{3D Room Segmentation}
Separately, several works have developed methods to segment
LiDAR reconstructions or floor plans into individual enclosed
spaces. 
Turner and Zakhor~\cite{turner2014floor} designed a room-segmentation pipeline 
that partitioned 2D point cloud maps into 2.5D building models via triangulation. 
To achieve 3D reconstruction, the approach assumed that interior walls 
were vertical and planar. 

More sophisticated methods were developed to 
parse 3D point clouds into rooms, 
such as detecting void spaces between walls 
using point density histograms~\cite{armeni20163d},
and extracting planar features before partitioning separate rooms 
via a multi-label energy minimisation formulation~\cite{mura2014automatic, mura2016piecewise}. 
These methods were limited to single-storey reconstructions. 

Ochmann~\etal~\cite{ochmann2019automatic} and 
Nikoohemat~\etal~\cite{Nikoohemat2020Routing} proposed methods that 
handle unstructured 3D point clouds for multi-storey room segmentation.
Ochmann~\etal introduced a versatile integer linear programming method to incorporate hard
constraints, e.g. wall connectivity, to ensure a plausible reconstruction.
Nikoohemat~\etal employed a mobile LiDAR SLAM system and separated building levels by
assuming that sloped trajectory segments represent staircase traversals. They
further segmented rooms using an
adjacency graph of planer segments.

While the effect of partitioning confined spaces is similar to our proposed system, 
we focus on online segmentation running onboard a mobile robot
instead of segmenting a complete reconstruction offline. 
Our proposed system aims to create segmented 3D reconstructions of 
confined spaces such as individual rooms and staircases on the fly
in the context of multi-floor exploration. 

\section{Methods}
\label{sec:Methods}

Our system expands upon the framework of~\cite{wang2020elastic} by
adding principled strategies which use 
spatial understanding 
and pose graph relative uncertainty,
to improve
the spawning and fusion of submaps in \textit{Graph Clustering}. 
\figref{fig:system_overview} provides an overview of the proposed system. 

An input to the reconstruction system is a pose graph published by a SLAM system~\cite{Ramezani2020LiDARSLAM}, 
made up of $Q+1$ nodes $X_k, k\in\{0,\dots,Q\}$
in a \textit{Registered Cloud List}. 
Each node describes the estimated pose of a LiDAR sensor frame 
$\{\mathcal{L}_k\}$ with respect to a fixed map frame $\{\mathcal{M}\}$, 
denoted as $^\mathcal{M}\mathbf{T}_{\mathcal{L}_k}\in\mathbf{SE}(3)$. 
Each node is also associated with a raw point cloud 
$C_k$ expressed in the LiDAR frame $\{\mathcal{L}_k\}$. 

The output of the reconstruction system consists of $N+1$
submaps. Each submap $\mathcal{S}_i, i\in\{0\text{...}N\}$
contains an accumulated submap cloud $C_{\mathcal{S}_i}$ in $\{\mathcal{M}\}$, 
a volumetric occupancy reconstruction in the submap frame $\{\mathcal{S}_i\}$
and its root pose $^\mathcal{M}\mathbf{T}_{\mathcal{S}_i}$. 

We compute and maintain Axis-Aligned Bounding Boxes (AABB) for each submap
in $\{\mathcal{M}\}$
using the reconstruction. 
Submap AABB can be significantly affected
by the orientation of $\{\mathcal{M}\}$, so we 
use AABBs not to 
accurately estimate scanned space, 
but as a lightweight method to determine
non-overlapping submaps (\secref{sec:Fusion}).

In the proposed system, we introduce the following set of measures
which we use to decide when to spawn or fuse submaps: 
\begin{itemize}
    \item \textit{Cloud Overlap Estimate}: A criteria to adjust the submap spawning decisions made by Local Mapping. 
    \item \textit{Submap Overlap Estimate}: A criteria to propose submap fusion for Global Mapping. 
    \item \textit{Relative Uncertainty}: A criteria to reject unreliable submap fusions. 
\end{itemize}
These measures will be explained in the following sections. 

\subsection{Cloud Overlap Estimate}
\label{sec:Spawning}

Many systems such as~\cite{reijgwart2020voxgraph} 
and~\cite{ho2018virtual} spawn submaps at fixed frequencies 
to bound the size of each submap as well as robot odometry drift within the submap. 
Our previous system~\cite{wang2020elastic} 
used a travel distance threshold to achieve the same purpose, 
under the assumption that the odometry drift 
is proportional to the distance travelled.

\begin{figure}[t]
    \centering
    \begin{subfigure}{\linewidth}
        \centering
        \includegraphics[width=0.9\linewidth,trim={0cm 1cm 0cm
            5cm},clip]{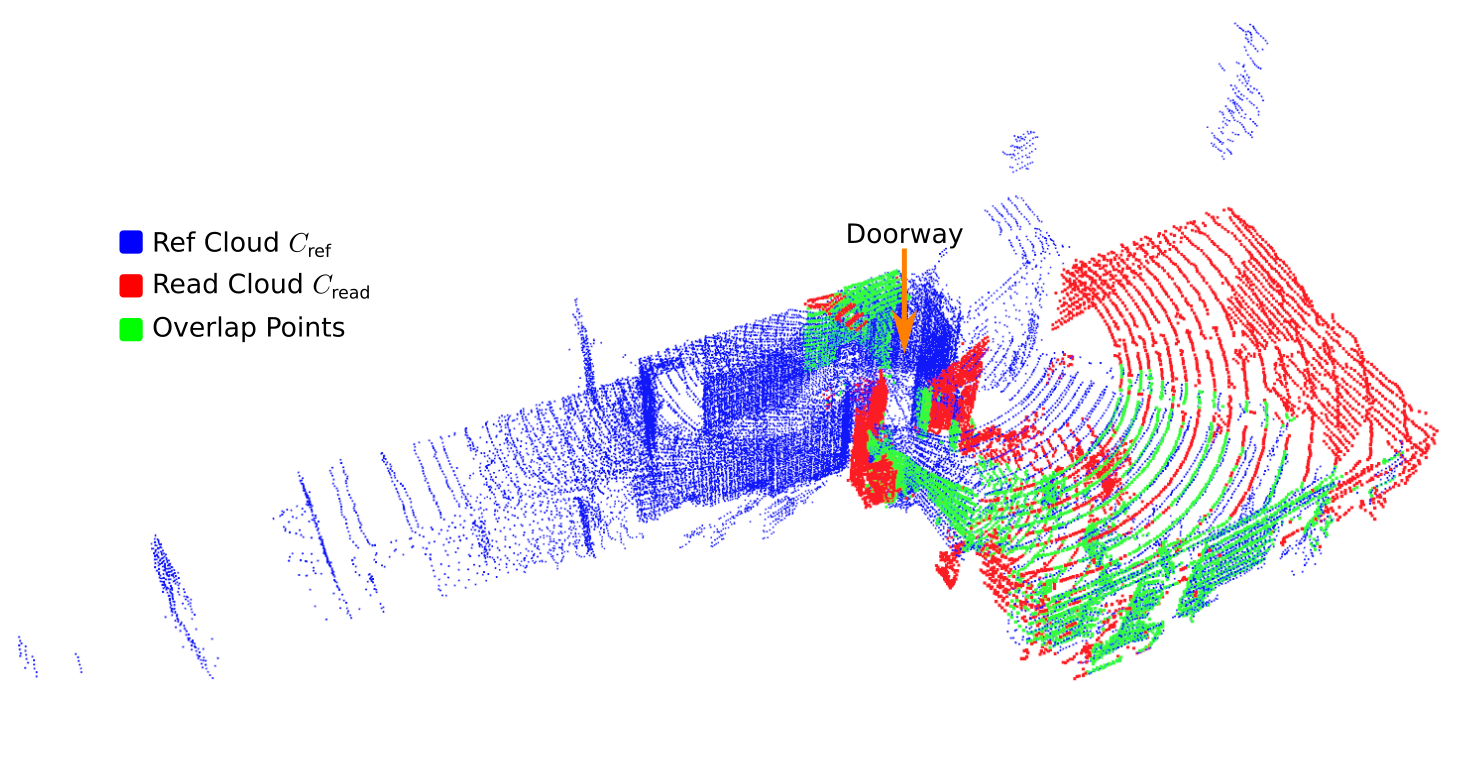}
    \end{subfigure}
    \begin{subfigure}{\linewidth}
        \centering
        \includegraphics[width=0.9\linewidth,trim={0cm 1cm 0cm
            5cm},clip]{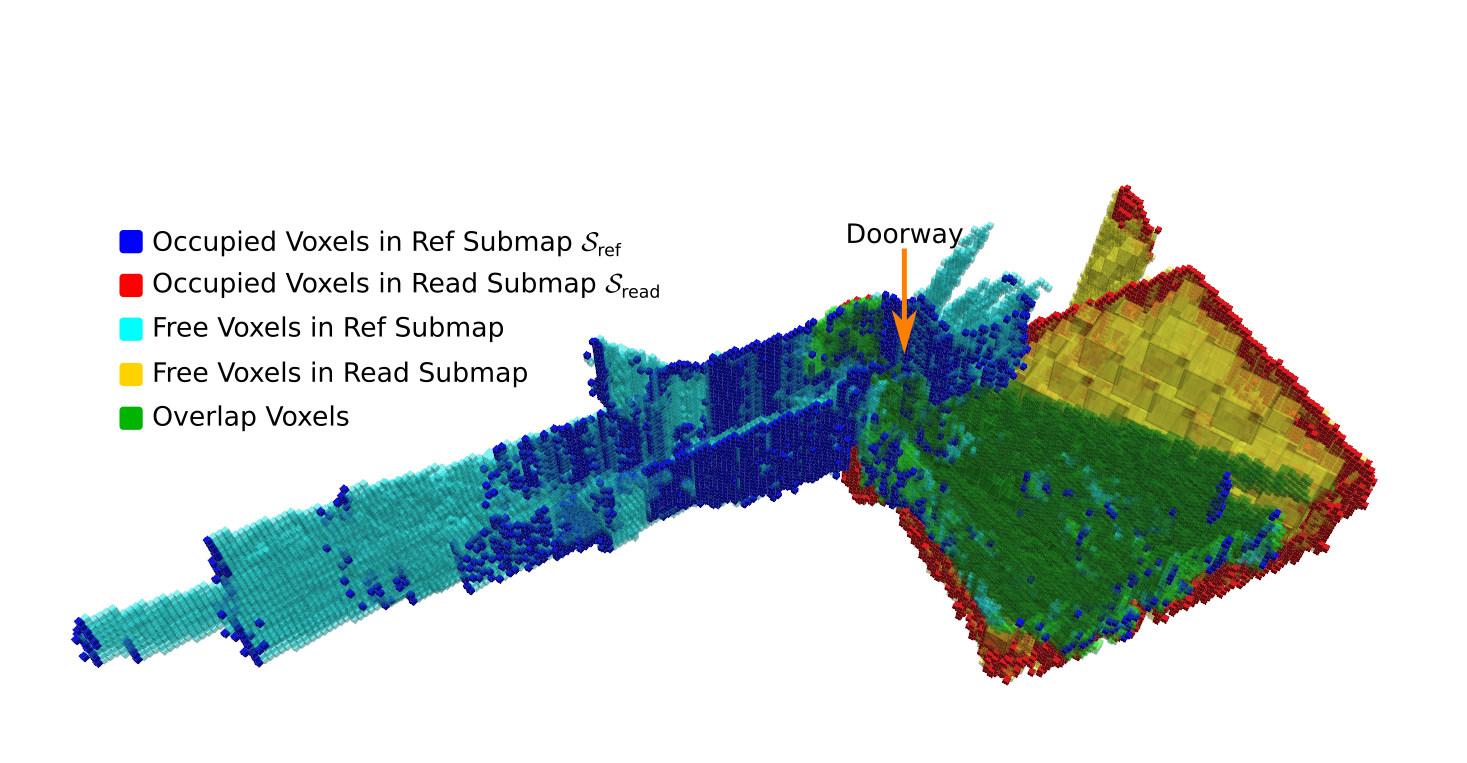}
    \end{subfigure}
    \caption{\small{An example of Cloud Overlap Estimate and Submap Overlap Estimate. \textbf{Top}: Limited overlap between the reference cloud $C_\text{ref}$ and the read cloud $C_\text{read}$ when entering a room through a narrow doorway. 
    \textbf{Bottom}: The voxel overlap between the volumetric occupancy reconstructions of the reference submap $\mathcal{S}_\text{ref}$ and the read submap $\mathcal{S}_\text{read}$. 
    }}
    \label{fig:overlap}
        \vspace{-5mm}
\end{figure}

Cloud Overlap Estimate adds another trigger 
to spawn submaps based on point cloud overlap. 
Because our odometry system is based on Iterative Closest Point 
(ICP)~\cite{Pomerleau12comp}, 
performance is affected by major changes 
in overlap such as when entering a new room~\cite{Simona2018Overlap}. 
Hence when traversing between two disconnected spaces via a narrow passage,
i.e. the scenario presented in~\figref{fig:overlap} (top),
it is beneficial to spawn a new submap and 
create an elastic connection. 
In a room network, our proposed system will spawn 
a new submap when going through 
a doorway and segment rooms online,
as demonstrated in~\figref{fig:front_page_full_map}.

\begin{algorithm}[t]
    \caption{ \small{Cloud Overlap Estimate.}}
    \label{alg:cloud_overlap}
    \small{
        \SetAlgoLined
        \DontPrintSemicolon
        \textbf{input}: New LiDAR cloud $C_\text{read}$ and submap cloud $C_{\mathcal{S}_\text{ref}}$, \;
        \textbf{output}: Cloud overlap ratio $R_\text{point,read}$\;
        \Begin{
            Voxel filter $C_\text{read}$ and $C_{\mathcal{S}_\text{ref}}$ to resolution $r_\text{filter}$ \;
            \For {\textup{Point} $P_i \subset C_\textup{read}$}
            {
                Search for $P_\text{neighbour} \subset C_{\mathcal{S}_\text{ref}}$ that is the closest to $P_i$\;
                \If{$\abs{\abs{P_i, P_\textup{neighbour}}} < \sqrt{3}\times r_\textup{filter}$}
                {
                    $N_\text{point,overlap} = N_\text{point,overlap} + 1$
                }
            }
            $R_\text{point,read} = N_{\text{point},\text{overlap}}/N_{\text{point},\text{read}}$\;
            return $R_\text{point,read}$
    }}
\end{algorithm}

\begin{figure*}[t]
    \centering
    \includegraphics[width=\linewidth,trim={0cm 0cm 0cm 0cm},clip]{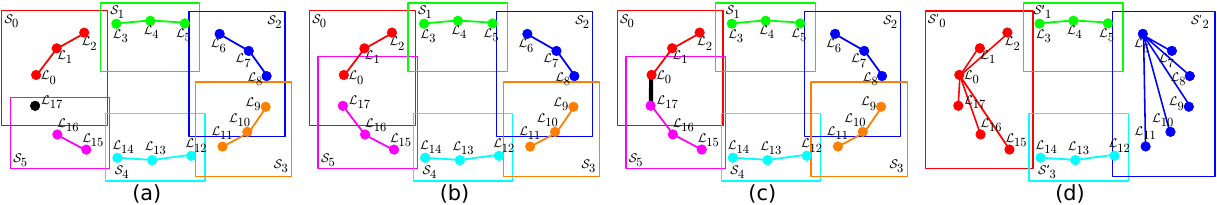}
    \caption{\small{An example of scan integration and submap fusion. 
            \textbf{(a)} -- There are $17$ LiDAR scans $\mathcal{L}_{0:16}$ in the existing reconstruction, clustered into $6$ submaps $\mathcal{S}_{0:5}$. 
            Each rectangle represents the AABB of each submap. 
            $\mathcal{L}_{17}$ is the latest new LiDAR scan, 
            and there have been no loop closures. 
            \textbf{(b)} -- \textit{Graph Clustering} allocates $\mathcal{L}_{17}$ to $\mathcal{S}_5$, 
            and the new scan passes the \textit{Cloud Overlap Estimate} criteria described in~\secref{sec:Spawning}. 
            Hence $\mathcal{L}_{17}$ is integrated into $\mathcal{S}_5$, expanding the AABB. 
            \textbf{(c)} -- There is a new loop closure edge (black line) given by the SLAM pose graph between $\mathcal{L}_0$ and $\mathcal{L}_{17}$. 
            The head and tail of the loop closure connection defines the submap overlap search range from $\mathcal{S}_0$ to $\mathcal{S}_5$. 
            \textbf{(d)} -- \textit{Graph Clustering} proposes the fusion of submaps $\mathcal{S}_5$ and $\mathcal{S}_0$ into $\mathcal{S}'_0$, and \textit{Submap Overlap Estimate} (\secref{sec:Fusion}) proposes the fusion between submaps $\mathcal{S}_3$ and $\mathcal{S}_2$ in $\mathcal{S}'_2$. 
            Both fusion proposals pass the \textit{Relative Uncertainty} criteria (\secref{sec:Uncertainty}) and are executed. 
            All submap indices are updated accordingly. 
            The AABBs of $\mathcal{S}_0$ and $\mathcal{S}_2$ are therefore expanded, 
            but their accumulated submap clouds are not, as explained in~\secref{sec:Spawning}. }}
    \label{fig:integration_and_fusion}
\end{figure*}

\algref{alg:cloud_overlap} presents how the overlap is measured
between the point cloud of a new scan $C_\text{read}$ and
the accumulated submap cloud $C_{\mathcal{S}_\text{ref}}$ 
of a reference submap $\mathcal{S}_\text{ref}$. 
Both $C_\text{read}$ and $C_{\mathcal{S}_\text{ref}}$ are
filtered to the same resolution $r_\text{filter}$
for uniformity in overlap estimation. 
If $C_\text{read}$ shares sufficient overlap 
($R_\text{point,read} > 0.6$)
with the accumulated submap cloud $C_{\mathcal{S}_\text{ref}}$, 
the new scan is integrated into a submap $\mathcal{S}_\text{ref}$, 
and the accumulated submap cloud $C_{\mathcal{S}_\text{ref}}$ grows
by adding $C_\text{read}$. 
For instance, in~\figref{fig:integration_and_fusion} (a) and (b), 
node $\mathcal{L}_{17}$ is integrated into submap $\mathcal{S}_5$, 
and LiDAR scan $C_{17}$ is accumulated into the submap cloud $C_{\mathcal{S}_5}$. 
However, we constrain point cloud accumulation of each submap
by not combining accumulated submap clouds together 
during submap fusion. 
In~\figref{fig:integration_and_fusion}, 
the volumetric occupancy submaps $\mathcal{S}_0$ and $\mathcal{S}_5$ in (c) 
are fused together into one submap $\mathcal{S}'_0$ in (d), 
but submap cloud $C_{\mathcal{S}_5}$ is not combined with $C_{\mathcal{S}_0}$. 
This stops these submap clouds from growing indefinitely 
as exploration continues. 

\subsection{Submap Overlap Estimate}
\label{sec:Fusion}

Submap fusion merges existing submaps, 
and reduces the memory usage of the overall system 
by fusing repeated reconstructions of the same physical space together.
Our previous pipeline~\cite{wang2020elastic} triggered
submap fusion using the loop closures detected in the SLAM system. 
Hence it only merged submaps that were created 
when the robot travelled very close to a previous pose.

In a large-scale (outdoor) environment, 
a long range ($\approx$\SI{60}{\meter}) LiDAR sensor  
can repeatedly scan the same space 
from poses that are far away from one another, 
resulting in significant redundancy between submaps
that loop closure fusion cannot address. 
Therefore we introduce an additional strategy for submap fusion 
based on the overlapping scanned spaces,
utilising the explicit representation of free space in 
our volumetric occupancy map.
This improves the reconstruction scalability when revisiting explored areas (although the SLAM pose
graph continues to grow linearly).

\begin{algorithm}[t]
    \caption{ \small{Submap Overlap Estimate.}}
    \label{alg:submap_overlap}
    \small{
        \SetAlgoLined
        \DontPrintSemicolon
        \textbf{input}: Pair of occupancy submaps $\mathcal{S}_\text{read}$ and $\mathcal{S}_\text{ref}$, \;
        \textbf{output}: Submap overlap ratios $R_\text{voxel,read}$ and $R_\text{voxel,ref}$\;
        \Begin{
            \For {\textup{Voxel} $V_\textup{read} \subset \mathcal{S}_\textup{read}$}
            {
                Find $V_\textup{ref} \subset \mathcal{S}_\text{ref}$ at the same coordinates as $V_\textup{read}$\;
                \If{$V_\textup{ref}$ is not unknown}
                {
                    \If{Both $V_\textup{ref}$ and $V_\textup{read}$ are free or occupied}
                    {
                        $N_\text{voxel,overlap} = N_\text{voxel,overlap} + 1$
                    }
                }
            }
            $R_\text{voxel,read} = N_{\text{voxel,overlap}}/N_{\text{voxel,read}}$\;
            $R_\text{voxel,ref} = N_{\text{voxel,overlap}}/N_{\text{voxel,ref}}$\;
            return $R_\text{voxel,read}$ and $R_\text{voxel,ref}$
    }}
\end{algorithm}

Submap overlap is computed by comparing the volumetric reconstruction 
as well as the occupancy information stored in 
each individual submap. \figref{fig:overlap} (bottom) demonstrates such a case.
\algref{alg:submap_overlap} describes the estimation of voxel overlap between a pair of submaps $\mathcal{S}_{\text{read}}$ and $\mathcal{S}_{\text{ref}}$. 
This pair of submaps are fused together 
if either the ratio $R_{\text{voxel},\text{ref}}$ or $R_{\text{voxel},\text{read}}$
exceed a configurable threshold $\lambda_{\text{fusion}}$. 
This threshold is set at $0.7$ to represent significant 
redundancy among submaps, and does not require further tuning between experiments. 

To ensure that the root poses of submaps are corrected
by loop closure before fusion, 
we define a submap overlap search range using the head and tail 
of each loop closure. 
For example, in~\figref{fig:integration_and_fusion} (c), 
SLAM loop closure is between $\mathcal{L}_0$ and $\mathcal{L}_{17}$,
and they belong to $\mathcal{S}_0$ and $\mathcal{S}_5$, respectively. 
Hence the search range for Submap Overlap Estimate is $\mathcal{S}_{0:5}$, 
and $\mathcal{S}_3$ and $\mathcal{S}_2$ are fused together 
due to significant submap overlap. 

Iterating through all voxels
is a computationally intense process. 
Therefore, we add a conservative but efficient 
preliminary heuristic based on the AABB of each submap
before computing submap voxel overlap.
Using the AABBs of
$\mathcal{S}_{\text{ref}}$ and $\mathcal{S}_{\text{read}}$,
we compute the volumetric overlap percentages
$\{R_{\text{aabb},\text{ref}},R_{\text{aabb},\text{read}}\}$ and 
compare them with $\lambda_{\text{fusion}}$. 
If both AABB overlaps are smaller than the threshold, 
such as $\mathcal{S}'_1$ and $\mathcal{S}'_4$ 
in~\figref{fig:integration_and_fusion} (c), 
the proposed system skips computing voxel overlap. 

\subsection{Relative Uncertainty}
\label{sec:Uncertainty}

In our current system design, 
after fusing a pair of submaps the individual submaps are discarded 
for memory efficiency. 
Each submap is internally rigid, so local consistency is essential. 
To improve submap fusion reliability and to retain global consistency, 
we propose a strategy that uses relative uncertainty 
between root poses of the submaps which are fused. 

The measurement of relative uncertainty 
allows us to detect and reject uncertain fusions. 
\figref{fig:integration_and_fusion} presents the case of fusing
two pairs of submaps, namely
$\mathcal{S}_0$ and $\mathcal{S}_5$, and $\mathcal{S}_2$ and $\mathcal{S}_3$. 
In the example of fusing $\mathcal{S}_0$ and $\mathcal{S}_5$, 
we first compute the relative uncertainty 
between the root poses of $\mathcal{S}_0$ and $\mathcal{S}_5$, 
which are $\mathcal{L}_0$ and $\mathcal{L}_{15}$. 

The proposed mathematical model for relative uncertainty 
between two poses is derived from the same notation as the GTSAM library~\cite{dellaert2017gtsam}, 
because we use it as the back-end of our pose graph SLAM system ~\cite{Ramezani2020LiDARSLAM}. 
See \appref{sec:relative-uncertainty} for the detailed derivation of the computation of relative uncertainty.

\begin{figure*}[t]
    \centering
    \includegraphics[width=\linewidth,trim={0cm 0cm 0cm 0cm},clip]{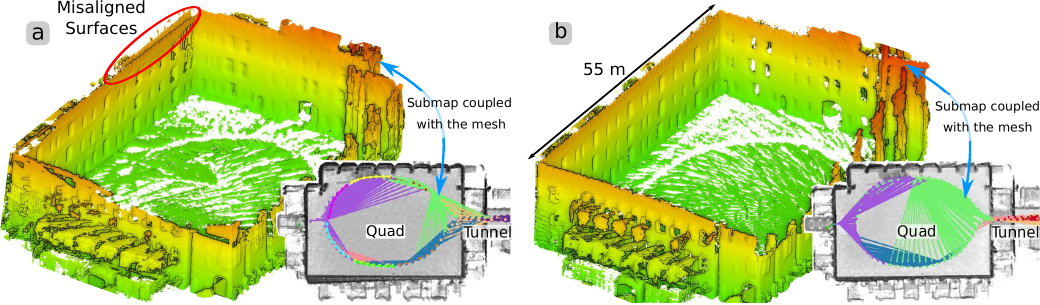}
    \caption{\small{
            The proposed spatial overlap analysis 
            improves the global consistency in the reconstruction. 
            Figure \textbf{(a)} and \textbf{(b)} presents the mesh and volumetric representations of NCD Long experiment, created by the baseline and the proposed system, respectively.  
            The volumetric representations are shown in grey overlaid with submap clusters.
            The mesh representations are created from the green submaps.
            Utilising spatial understanding of the environment leads to more reliable submap fusion and therefore better alignment. 
            The mesh created by the previous pipeline has misaligned double surfaces while the
proposed system improves the consistency in the mesh.}}
    \label{fig:ncd_accuracy}
\end{figure*}

We compute the eigenvalues of the relative uncertainty matrix as a quality metric of the fusion. 
For $\mathbf{SE}(3)$ transformations 
${^\mathcal{M}\mathbf{T}}_{\mathcal{S}_0}, {^\mathcal{M}\mathbf{T}}_{\mathcal{S}_5} \in \mathbb{R}^{6}$, 
the relative uncertainty $\mathbf{\Sigma}_{\mathcal{S}_0\mathcal{S}_5}$ 
is a $6 \times 6$ matrix, 
and there are $6$ eigenvalues --- $3$ for translation and $3$ for rotation. 
We compare the $3$ eigenvalues of translation against 
a configurable threshold $\lambda_{\text{uncertainty}}$. 
If one of the eigenvalues exceeds the threshold, 
the fusion between $\mathcal{S}_0$ and $\mathcal{S}_5$ is 
rejected. 
We set $\lambda_{\text{uncertainty}}$ to be \SI{0.2}{\meter}
in our experiments based on our SLAM noise model, 
which is not subject to change between experiments. 

\section{Experiments and Evaluation}
\label{sec:Experiments}

\begin{figure}[t]
    \centering
    \includegraphics[width=0.8\linewidth,trim={0cm 0cm 0cm 0cm},clip]{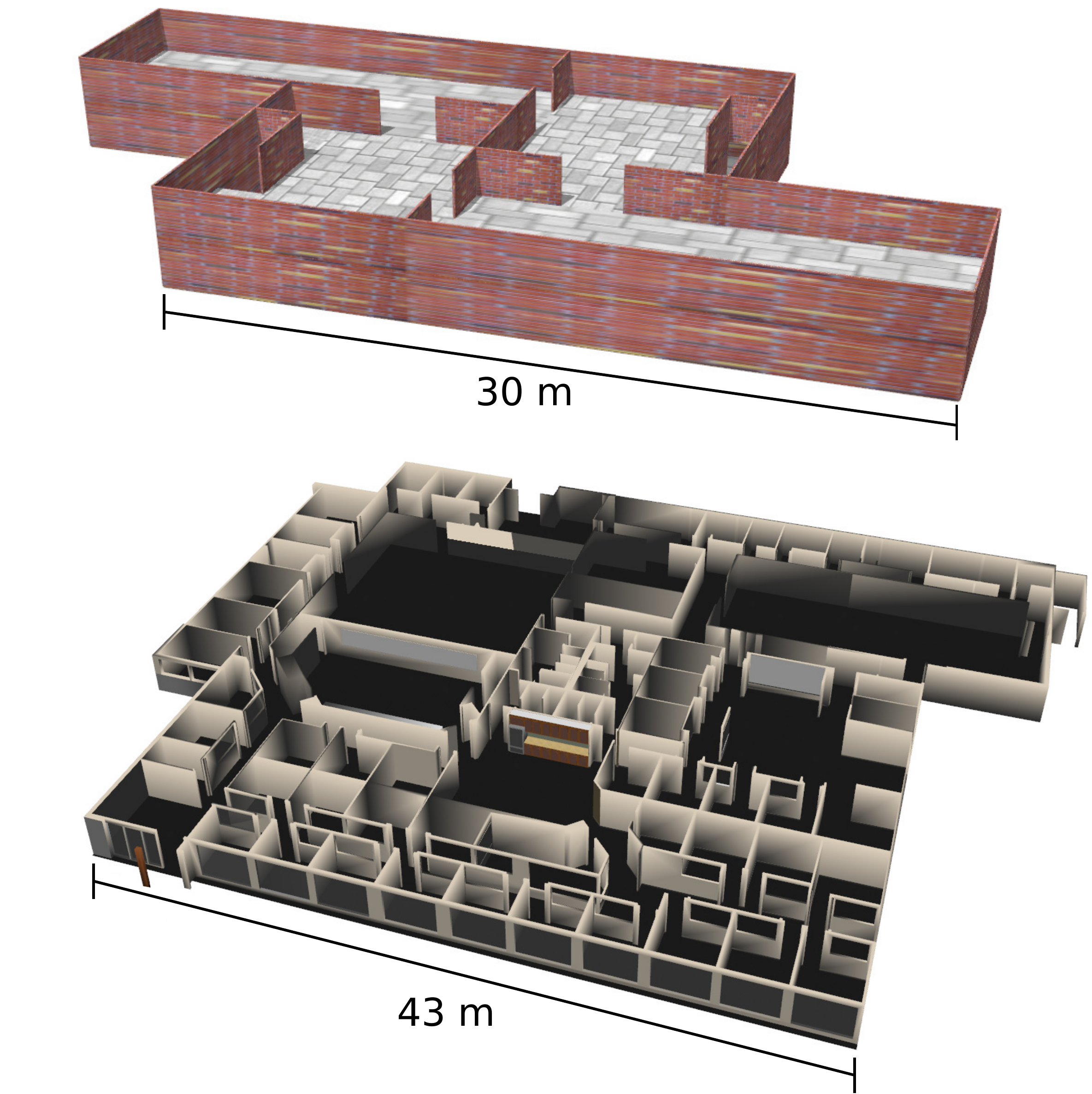}
    \caption{\small{
            The Gazebo environments of a small (top) and a large (bottom) room network for
experiments in simulation.}}
    \label{fig:sim_environment}
        \vspace{-3mm}
\end{figure}

In this section we evaluate the improvement in system scalability 
and global consistency brought about by the proposed 
submap spawning and fusion strategies, 
compared with our previous reconstruction pipeline~\cite{wang2020elastic} 
as the baseline. 
The proposed system has been assessed using 
a large-scale outdoor experiment 
with a handheld device 
in the Newer College Dataset (NCD Long experiment)~\cite{ramezani2020newer}
and a multi-storey multi-room exploration experiment with a Boston Dynamics Spot robot 
(ORI experiment), as shown in~\figref{fig:front_page_full_map}.
Lastly, we include experiments in a small and a large room network (\figref{fig:sim_environment})
using the Gazebo simulator to 
demonstrate the performance of the proposed system. 
Both experiments in simulation took on loopy trajectories, 
with the exploration in the large room network being longer
and more complex. 

\tabref{table:experiment_sensors} gives details of the different LiDAR sensors
used in these experiments. 
The LiDAR sensors produce organised point cloud scans of
$64\times1024$ points at \SI{10}{\hertz}. 
The SLAM system created a node in its pose graph every \SI{2}{\meter}
travelled when exploring.
The proposed system and the baseline integrated
the LiDAR scans at each SLAM node using 
the \textit{MultiresOFusion} mode in~\cite{wang2020elastic}
for volumetric occupancy reconstruction. 
The voxel resolution used in these experiments was \SI{6.5}{\centi\meter} 
and we integrated LiDAR ranges between \SI{0.5}{\meter} and \SI{60}{\meter}. 
These settings give high resolution and long range 
while retaining \SI{3}{\hertz} integration.

Surface mesh representations of the reconstruction, 
for example~\figref{fig:ncd_accuracy}, 
were created by applying the Marching Cubes algorithm~\cite{Lorensen1987MarchingCubes}
on the zero-crossings of the occupancy. 

\begin{table}
    \centering
    \resizebox{\columnwidth}{!}{%
        \begin{tabular}{|l|cccc|}
            \hline \cellcolor{orange!25}& \multicolumn{4}{c|}{\cellcolor{orange!25}LiDAR properties} \\  
            \cellcolor{orange!25}Experiment& \cellcolor{orange!25}Model & \cellcolor{orange!25}Vertical FoV  & \cellcolor{orange!25}Horizontal FoV & \cellcolor{orange!25}Max range~(m) \\ 
            \hline \hline 
            NCD Long (handheld) & Ouster OS1-64 & \SI{33.2}{\degree} & \SI{360}{\degree} & 120 \\ 
            \hline 
            ORI (with Spot) & Ouster OS0-64 & \SI{90}{\degree} & \SI{360}{\degree} & 50 \\ 
            \hline 
            Simulation (with UGV) & Ouster OS0-64 & \SI{90}{\degree} & \SI{360}{\degree} & 50 \\ 
            \hline
        \end{tabular}
    }
    \caption{LiDAR sensors used in the experiments and their properties. FoV: Field of View}
    \label{table:experiment_sensors}
    \vspace{-3mm}
\end{table}

\subsection{Large-scale Outdoor Experiment}

In this section we present the performance of the proposed system 
and the baseline when tested using NCD Long. 
This dataset consists of a \SI{2.2}{\kilo\meter} 
exploration over \SI{44}{\minute}
in a \SI{135x122}{\square\meter} environment. 
The principled clustering strategies in the proposed system 
improve the consistency of the global reconstruction, 
as shown in~\figref{fig:ncd_accuracy}. 
The proposed system also demonstrates improved scalability in 
memory usage compared with the baseline. 

\subsubsection{Reconstruction Accuracy}

\figref{fig:ncd_accuracy} (a) shows the mesh and 
volumetric reconstructions created by the baseline. 
In particular the green submap in the Quad area was created 
as a rigid fusion of several earlier submaps after multiple loop closures were established. 
As shown in the birds-eye view next
to the mesh representation, 
this submap contains scans taken both in the Quad and the Tunnel. 
These scans have limited overlap with one another
and registration between them is unreliable as a result. 
This led to a duplicate reconstruction of the indicated wall of the Quad. 

\begin{figure}[t]
    \centering
    \begin{subfigure}{\linewidth}
        \includegraphics[width=\linewidth,trim={0cm 0cm 0cm 0cm},clip]{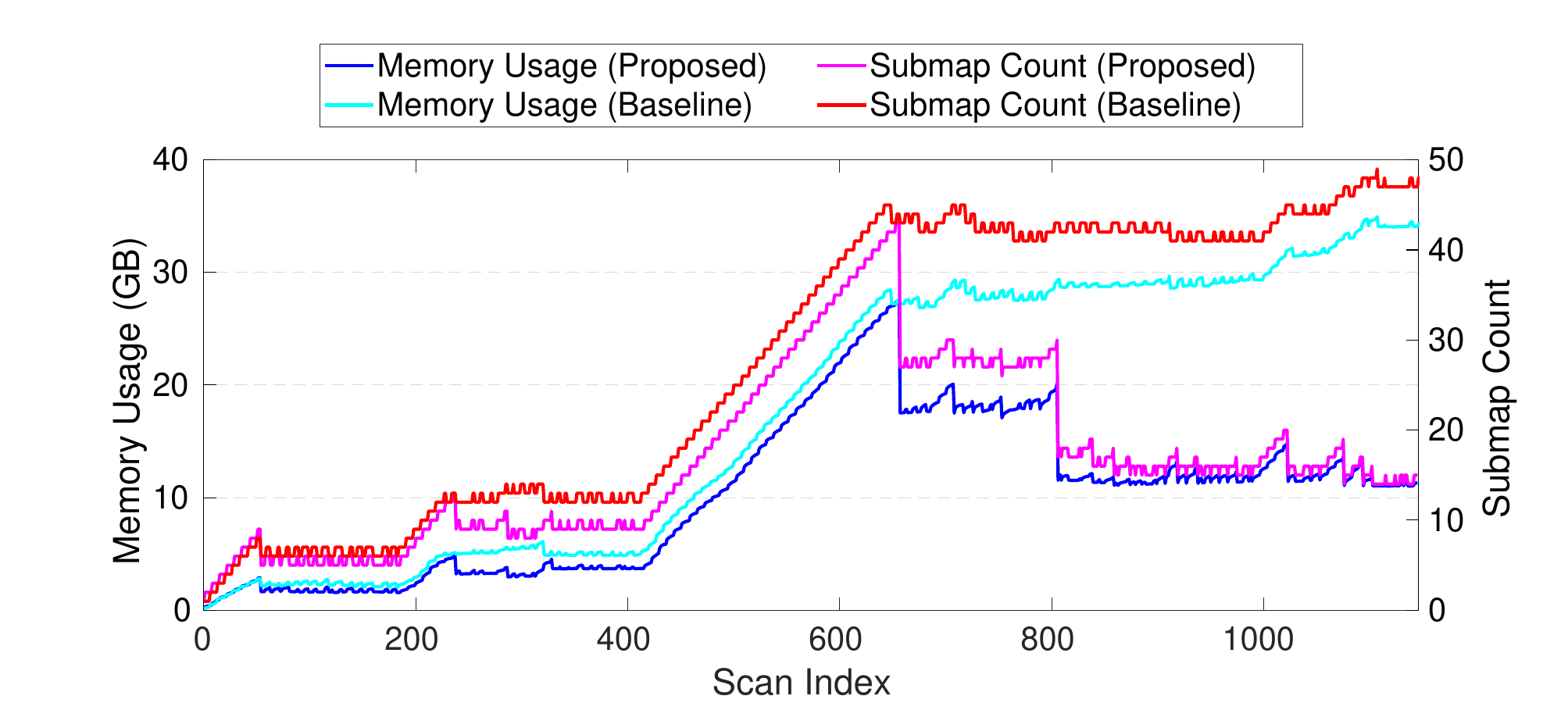}
        \caption{\small{The performance in the NCD Long experiment}}
        \label{fig:ncd_memory}
    \end{subfigure}
    \begin{subfigure}{\linewidth}
    \includegraphics[width=\linewidth,trim={0cm 0cm 0cm 0cm},clip]{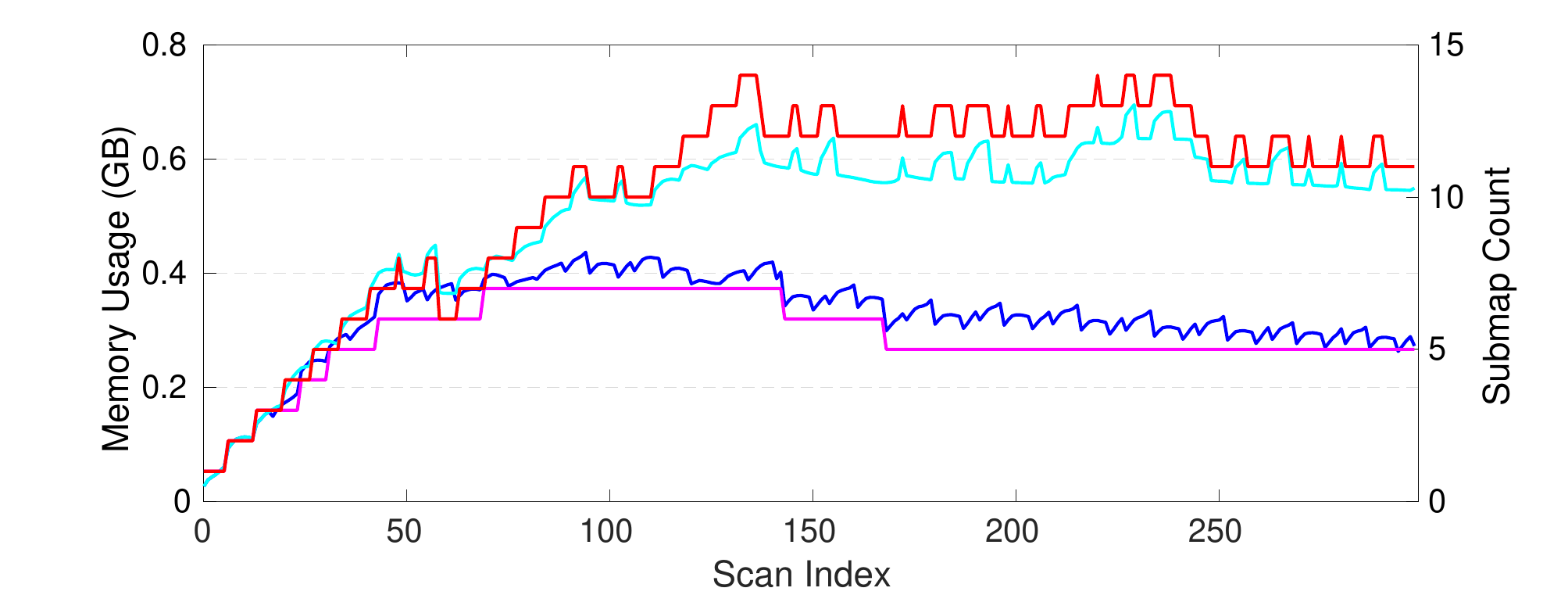}
    \caption{\small{The performance in the small room network (simulated)}}
    \label{fig:sim_memory}
    \end{subfigure}
    \begin{subfigure}{\linewidth}
    \includegraphics[width=\linewidth,trim={0cm 0cm 0cm 0cm},clip]{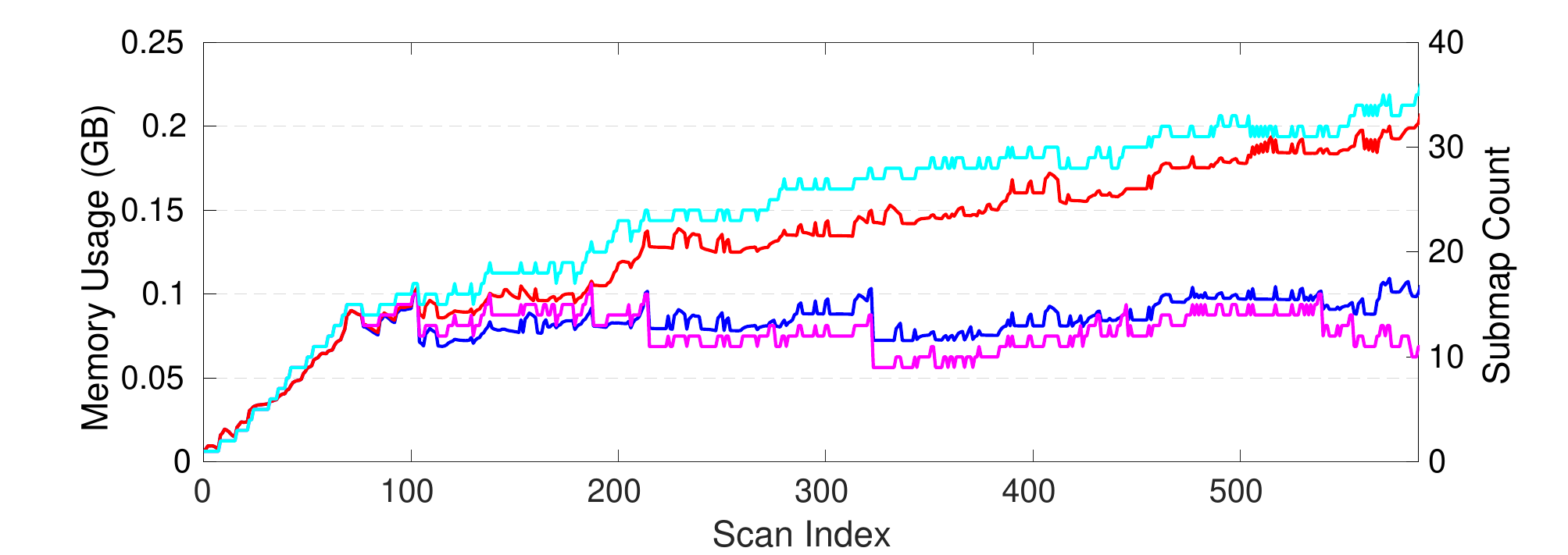}
    \caption{\small{The performance in the large room network (simulated)}}
    \label{fig:garage_memory}
    \end{subfigure}
    \caption{\small{The memory usage and submap counters of the proposed system and the baseline system.}}
    \label{fig:overlap_memory}
        \vspace{-4mm}
\end{figure}

By introducing our proposed submap fusion algorithms, the consistency of reconstruction 
is improved as shown in \figref{fig:ncd_accuracy} (b).
The Cloud Overlap Estimate strategy makes better decisions when
the handheld device travels between the Quad 
and the Tunnel --- maintaining elastic connections between these 
two spaces in the global reconstruction. 
The measurement of relative uncertainty also rejects
unreliable submap fusions. The global volumetric map
and the submap mesh both demonstrate improved accuracy in surface alignment. 

The two submap reconstructions presented in~\figref{fig:ncd_accuracy}
were also compared with the ground truth point cloud
provided in the dataset~\cite{ramezani2020newer}. 
We used CloudCompare\footnote{https://www.danielgm.net/cc/} to
sample dense point clouds from both meshes, 
align the reconstructed point clouds with the ground truth, 
and compute the point-to-point distance error between them. 
The reconstruction created by the proposed system (\figref{fig:ncd_accuracy} (b)) 
has about \SI{90}{\percent} points with error less than \SI{25}{\centi\meter}, 
while that created by the baseline (\figref{fig:ncd_accuracy} (a)) 
has only about \SI{75}{\percent}. 
\SI{90}{\percent} of the points in the baseline mesh has errors 
less than \SI{50}{\centi\meter}. 

\subsubsection{Memory Consumption}

In the proposed system, submaps that have significantly overlapping
scan volumes are fused according to their Submap Overlap Estimate.
This is in addition to submap fusion based on loop closures, 
and properly allows the map to scale
with the size of the environment 
rather than the length of the exploration. 
\figref{fig:overlap_memory} (a) presents
the memory usage
and submap counter of the proposed system 
compared to the baseline in NCD Long. 
We compute the memory usage by summing the size of 
allocated memory for each submap's octree in RAM. 

In our previous work~\cite{wang2020elastic}, 
the baseline method was compared against 2 state-of-the-art
reconstruction pipelines, 
i.e. OctoMap~\cite{hornung2013octomap} and Voxgraph~\cite{reijgwart2020voxgraph}, 
and demonstrated its advantage of efficient reconstruction memory usage. 

In the baseline method the submaps can only be merged when
loop closures occur which causes memory usage to grow over time. In contrast,
our proposed approach can merge spatially overlapping volumes such that the number of submaps
can plateau. For the NCD Long experiment, submap count stabilised at
about $30$ submaps when the entire environment has been explored at scan $650$.
Memory usage actually decreased after scan $650$ to
\SI{\sim18}{\giga\byte} while maintaining the \SI{6.5}{\centi\meter} resolution reconstruction.
By the end of the experiment, there was a \SI{65}{\percent} reduction
in memory usage compared to the baseline.

\begin{figure}[t]
    \centering
    \begin{subfigure}{0.9\linewidth}
        \includegraphics[width=0.9\linewidth,trim={0cm 0cm 0cm 0cm},clip]{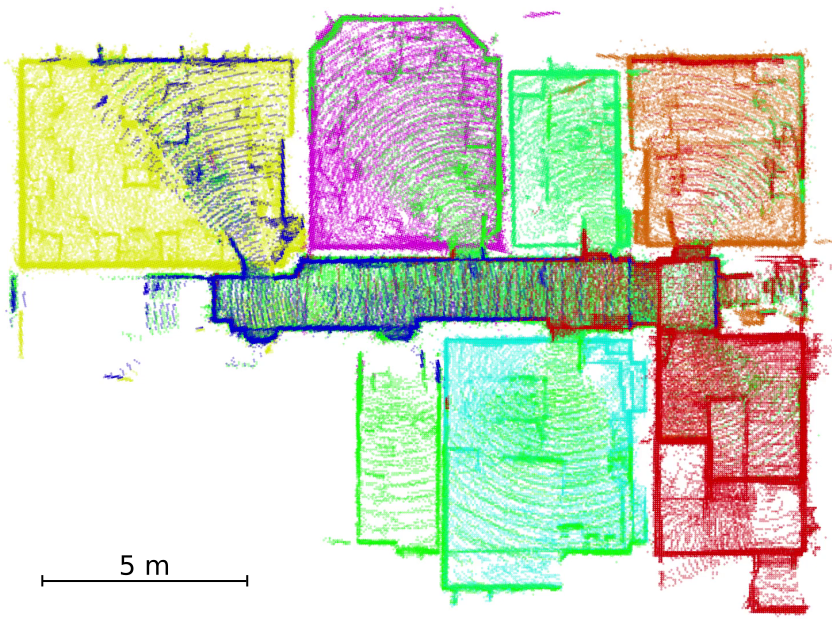}
    \end{subfigure}
    \begin{subfigure}{0.9\linewidth}
        \includegraphics[width=0.9\linewidth,trim={0cm 0cm 0cm
            0cm},clip]{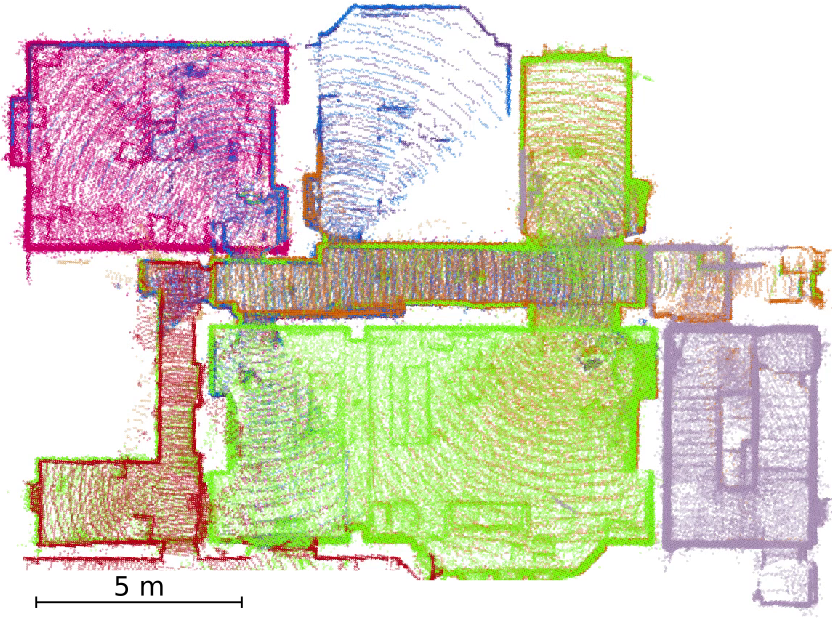}
    \end{subfigure}
    \begin{subfigure}{0.9\linewidth}
        \includegraphics[width=0.9\linewidth,trim={0cm 0cm 0cm
            0cm},clip]{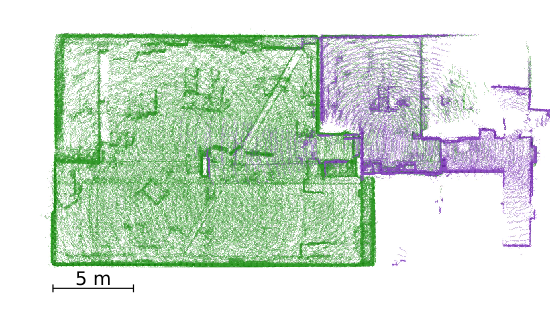}
    \end{subfigure}
    \caption{\small{The volumetric submap reconstructions of each floor (\textbf{Top:} first; \textbf{Middle:} ground; \textbf{Bottom:} basement) in ORI, with each room segmented into unique submaps by the proposed system on the fly during exploration. }}
    \label{fig:ori_segmentation}
        \vspace{-4mm}
\end{figure}

\subsection{Multi-storey Multi-room Indoor Exploration}

In the ORI experiment a quadruped robot Spot explored
three floors of a typical university research lab (\figref{fig:front_page_full_map}); 
we show the reconstruction of every floor in the building in \figref{fig:ori_segmentation}. 
New submaps were spawned when the robot entered or exited rooms
because the proposed system detected 
a decrease in cloud overlap. 
Spatial overlap analysis then merged overlapping submaps in each room, 
creating a unique reconstruction for each enclosed space. 
It further ensured that these submaps remained independent
allowing future SLAM loop closures to re-position the room submaps as needed. 
In the supplementary video we demonstrate the full experiment 
and the incremental mapping of the building. 

Segmenting rooms on the fly allows real-time applications
such as path planning and obstacle avoidance to consider only the minimal submaps rather than the entire global reconstruction. 
The proposed system can thus improve the scalability of other
applications.

\begin{figure}[t]
    \centering
    \begin{subfigure}{\linewidth}
        \includegraphics[width=\linewidth,trim={0cm 0cm 0cm
            0cm},clip]{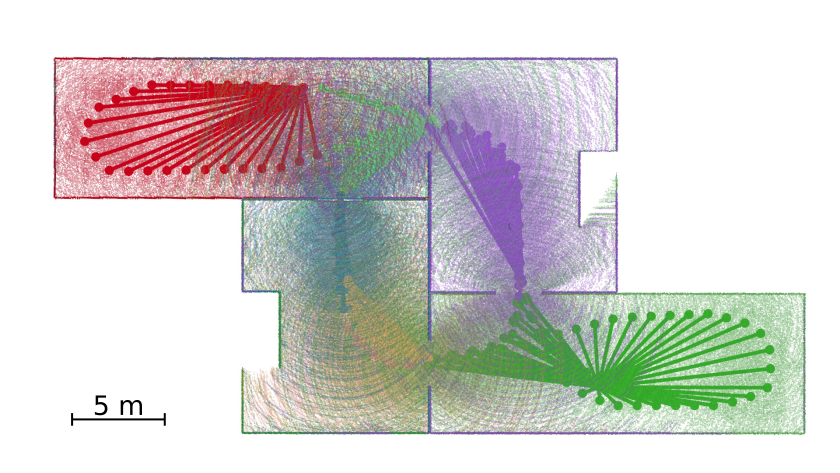}
        \caption{\small{Submap reconstruction in the small room network}}
        \label{fig:sim_segmentation}
    \end{subfigure}
    \begin{subfigure}{0.9\linewidth}
        \includegraphics[width=\linewidth,trim={0cm 0cm 0cm
            0cm},clip]{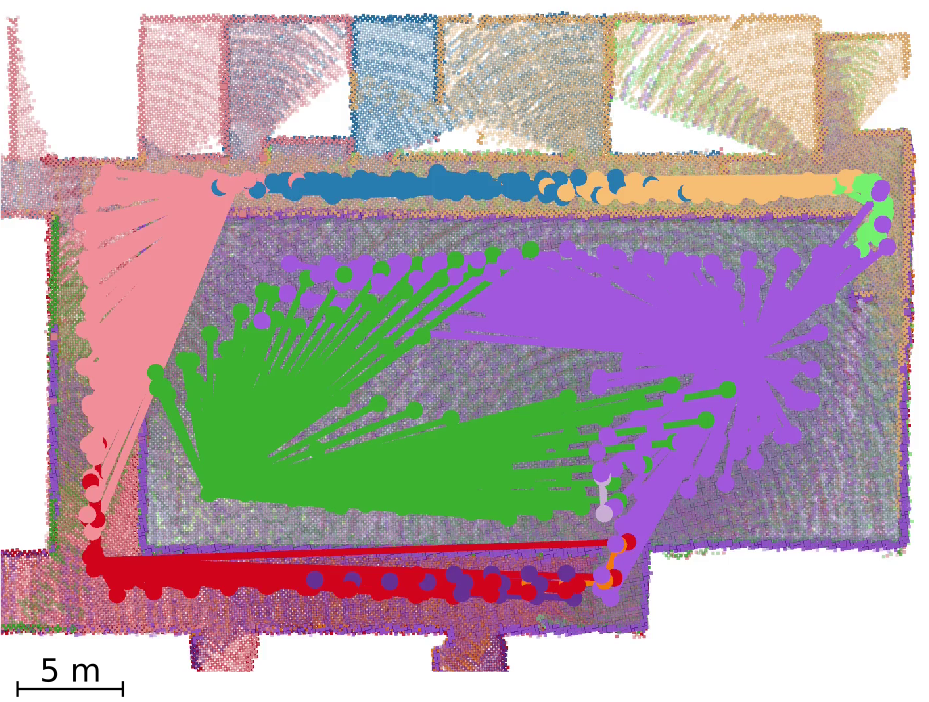}
        \caption{\small{Submap reconstruction in the large room network}}
        \label{fig:garage_segmentation}
    \end{subfigure}
    \caption{\small{The proposed system segments individual rooms and fuses redundant submaps during simulation experiments. }}
    \label{fig:room_segmentation}
        \vspace{-4mm}
\end{figure}

\subsection{Room Networks in Simulation}

To test the performance for long term operation missions, 
we carried out 
two experiments in Gazebo simulation with a wheel robot carrying a simulated 3D LiDAR in the environments in~\figref{fig:sim_environment}.

In the small room network, 
the mission looped around the environment three times 
in both directions. 
As shown by the clustered poses in~\figref{fig:room_segmentation} (a) 
there is a clear division between submaps at each doorway. 
Each room was constructed with either one or two submaps ---
even after multiple revisits. 
\figref{fig:room_segmentation} (b) demonstrated 
an approximately \SI{700}{\meter} exploration 
in a section of the large room network. 
Similarly, the room in the middle only contains 
two major submaps even after about $10$ loops with different routes. 
Scans along corridors were also fused together 
based on overlap and loop closures, creating only $10$
submaps in the end. 

\figref{fig:overlap_memory} (b) and (c) present
the memory usage in both simulation experiments, 
where the submap count and memory usage plateaued 
as the environments were repeatedly scanned. 
Overall, the proposed method decreased the memory usage 
of both experiments by \SI{50}{\percent} compared to the baseline method 
even after extensive exploration and revisiting, 
such as the experiment in the large room network (\figref{fig:room_segmentation} (b)). 


\section{Conclusion and Future Work}
\label{sec:FutureWork}

This paper introduced a set of principled
strategies that leverage spatial overlap analysis
to improve the spawning and fusion of 3D reconstruction submaps. 
The proposed system can merge together submaps
that scan the same space,
allowing the memory consumption of our reconstruction 
to scale with the volume of the explored area rather than the duration of operation.
Additionally, we improved submap fusion reliability by directly computing
relative uncertainty between pairs of poses in the SLAM pose graph.
These strategies delay submap fusion
in 3D reconstruction until there is sufficient confidence
to improve global consistency of the overall map. 

Furthermore, our system can segment individual rooms on the fly
as a robot explores a room network by using spatial overlap. 
This feature can be beneficial to planners exploring room networks 
such as in the DARPA SubT challenge. 

In future work, we would like to introduce semantic information into the reconstruction, 
further segmenting objects of interest from the scene for tasks such as industrial inspections. 
We also aim to design a system that can autonomously infer floor plans from a multi-storey 
reconstruction, which enables high-level path planning between floors. 

\appendices
\section{Computation of Relative Uncertainty in Pose Graph SLAM}
\label{sec:relative-uncertainty}
To compute the relative uncertainty between any pair of nodes in our pose graph,
we derived the following formula
based on the conventions of GTSAM
with reference to
Mangelson~\etal\cite{mangelson2020characterizing}.

First we define the probability distribution of the relative transformation
from submap $\mathcal{S}_i$ to node $\mathcal{S}_j$ as:
\begin{equation}
\label{relative-t}
^{\mathcal{S}_i}\mathbf{T}_{\mathcal{S}_j} =
{^\mathcal{M}\mathbf{T}_{\mathcal{S}_i}^{-1}}{^\mathcal{M}\mathbf{T}_{\mathcal{S}_j}}
\end{equation}
where the poses ${^{\mathcal{M}}}\mathbf{T}_{\mathcal{S}_i}$ and
${^{\mathcal{M}}}\mathbf{T}_{\mathcal{S}_j}$ indicate
probability
distributions on $\mathbf{SE}(3)$ following a right-hand composition:
\begin{equation}
\label{gtsam}
\mathbf{T} = \bar{\mathbf{T}}\ \text{Exp}(\xi)
\end{equation}
$\bar{\mathbf{T}}$ is the mean transformation of the distribution, and $\xi$ is
a
perturbation that follows a Gaussian distribution. In \eqref{relative-t} we
consider that the poses have covariances
$\Sigma_{\mathcal{MS}_i}$ and $\Sigma_{\mathcal{MS}_j}$, respectively.

In order to derive the expressions for the relative uncertainty, we need the
adjoint action of $\bar{\mathbf{T}}$ on $\xi$, denoted as
$\text{Ad}_{\bar{\mathbf{T}}}(\xi)$, defined as
follows:
\begin{equation}
\begin{split}
\label{adjoint}
\text{Ad}_{\bar{\mathbf{T}}}(\xi) := \text{Ad}_{\bar{\mathbf{T}}}\xi &=
\text{Log}({\bar{\mathbf{T}}}\ \text{Exp}(\xi){\bar{\mathbf{T}}}^{-1})\\
\text{Exp}(\text{Ad}_{\bar{\mathbf{T}}}\xi) &=
{\bar{\mathbf{T}}}\ \text{Exp}(\xi){\bar{\mathbf{T}}}^{-1} \\
{\bar{\mathbf{T}}}^{-1}\text{Exp}(\text{Ad}_{\bar{\mathbf{T}}}\xi) &=
\text{Exp}(\xi){\bar{\mathbf{T}}}^{-1} \\
\text{Exp}(\xi){\bar{\mathbf{T}}} &= {\bar{\mathbf{T}}}\
\text{Exp}(\text{Ad}_{{\bar{\mathbf{T}}}^{-1}}\xi)
\end{split}
\end{equation}

Expanding~\eqref{relative-t} using \eqref{gtsam} and~\eqref{adjoint}:
\begin{equation}
\begin{split}
\label{golden-eq}
^{\mathcal{S}_i}{\bar{\mathbf{T}}}_{\mathcal{S}_j} &
\text{Exp}(^{\mathcal{S}_i}\xi_{\mathcal{S}_j}) \\
=\ &
\text{Exp}(-^{\mathcal{M}}\xi_{\mathcal{S}_i})
{^{\mathcal{M}}{\bar{\mathbf{T}}}_{\mathcal{S}_i}^{-1}}
{^{\mathcal{M}}{\bar{\mathbf{T}}}_{\mathcal{S}_j}}
\text{Exp}(^{\mathcal{M}}\xi_{\mathcal{S}_j})\\
=\ &
{^{\mathcal{M}}{\bar{\mathbf{T}}}_{\mathcal{S}_i}^{-1}}
\text{Exp}(-\text{Ad}_{^{\mathcal{M}}{\bar{\mathbf{T}}}_{\mathcal{S}_i}}
{^{\mathcal{M}}\xi_{\mathcal{S}_i}})
{^{\mathcal{M}}{\bar{\mathbf{T}}}_{\mathcal{S}_j}}
\text{Exp}(^{\mathcal{M}}\xi_{\mathcal{S}_j})\\
=\ &
{^{\mathcal{M}}{\bar{\mathbf{T}}}_{\mathcal{S}_i}^{-1}}
{^{\mathcal{M}}{\bar{\mathbf{T}}}_{\mathcal{S}_j}}
\text{Exp}(-\text{Ad}_{{^{\mathcal{M}}{\bar{\mathbf{T}}}_{\mathcal{S}_j}^{-1}}}
\text{Ad}_{^{\mathcal{M}}{\bar{\mathbf{T}}}_{\mathcal{S}_i}}
{^{\mathcal{M}}\xi_{\mathcal{S}_i}})
\text{Exp}(^{\mathcal{M}}\xi_{\mathcal{S}_j})
\end{split}
\end{equation}

Let $^{\mathcal{S}_i}{\bar{\mathbf{T}}}_{\mathcal{S}_j} \triangleq
{^\mathcal{M}{\bar{\mathbf{T}}}_{\mathcal{S}_i}^{-1}}{^\mathcal{M}{\bar{\mathbf{T}}}_{\mathcal{S}_j}
}$,
 then we can establish the following equivalence:
\begin{equation}
\text{Exp}(^{\mathcal{S}_i}\xi_{\mathcal{S}_j}) =
\text{Exp}(-\text{Ad}_{{^{\mathcal{M}}{\bar{\mathbf{T}}}_{\mathcal{S}_j}^{-1}}}
\text{Ad}_{^{\mathcal{M}}{\bar{\mathbf{T}}}_{\mathcal{S}_i}}
{^{\mathcal{M}}\xi_{\mathcal{S}_i}})
\text{Exp}(^{\mathcal{M}}\xi_{\mathcal{S}_j})
\end{equation}

The covariance of the perturbation on the left should be equal to the one on
the right. However, we cannot compute the covariance directly because of
the properties of the exponential map. Instead, we define
$^{\mathcal{S}_i}\xi_{\mathcal{S}_j}'=-\text{Ad}_{{^{\mathcal{M}}{\bar{\mathbf{T}}}_{\mathcal{S}_j}^
{-1}}}
\text{Ad}_{^{\mathcal{M}}{\bar{\mathbf{T}}}_{\mathcal{S}_i}}
{^{\mathcal{M}}\xi_{\mathcal{S}_i}}$, and use the Baker-Campbell-Hausdorff
(BCH) formula~\cite{klarsfeld1989baker} up to first order:
\begin{equation}
\begin{split}
E[{^{\mathcal{S}_i}\xi_{\mathcal{S}_j}}{^{\mathcal{S}_i}\xi_{\mathcal{S}_j}^T}]
& \approx
E[{^{\mathcal{M}}\xi_{\mathcal{S}_i}'}{^{\mathcal{M}}\xi_{\mathcal{S}_i}'^T}] +
E[{^{\mathcal{M}}\xi_{\mathcal{S}_j}}{^{\mathcal{M}}\xi_{\mathcal{S}_j}^T}] \\
& + E[{^{\mathcal{M}}\xi_{\mathcal{S}_i}'}{^{\mathcal{M}}\xi_{\mathcal{S}_j}^T}] +
E[{^{\mathcal{M}}\xi_{\mathcal{S}_j}}{^{\mathcal{M}}\xi_{\mathcal{S}_i}'^T}]
\end{split}
\end{equation}
which, after computing the covariance terms, provides an approximation for the
covariance of the relative transformation:
\begin{equation}
\begin{split}
\mathbf{\Sigma}_{\mathcal{S}_i\mathcal{S}_j} &\approx
(\text{Ad}_{{^{\mathcal{M}}{\bar{\mathbf{T}}}_{\mathcal{S}_j}^{-1}}}
\text{Ad}_{^{\mathcal{M}}{\bar{\mathbf{T}}}_{\mathcal{S}_i}})
\mathbf{\Sigma}_{\mathcal{M}\mathcal{S}_i}
(\text{Ad}_{{^{\mathcal{M}}{\bar{\mathbf{T}}}_{\mathcal{S}_j}^{-1}}}
\text{Ad}_{^{\mathcal{M}}{\bar{\mathbf{T}}}_{\mathcal{S}_i}})^T \\
&
+ \mathbf{\Sigma}_{\mathcal{M}\mathcal{S}_j} \\
&
- (\text{Ad}_{{^{\mathcal{M}}{\bar{\mathbf{T}}}_{\mathcal{S}_j}^{-1}}}
\text{Ad}_{^{\mathcal{M}}{\bar{\mathbf{T}}}_{\mathcal{S}_i}})
\mathbf{\Sigma}_{\mathcal{M}\mathcal{S}_i,\mathcal{M}\mathcal{S}_j} \\
&
- \mathbf{\Sigma}_{\mathcal{M}\mathcal{S}_i,\mathcal{M}\mathcal{S}_j}^T
(\text{Ad}_{{^{\mathcal{M}}{\bar{\mathbf{T}}}_{\mathcal{S}_j}^{-1}}}
\text{Ad}_{^{\mathcal{M}}{\bar{\mathbf{T}}}_{\mathcal{S}_i}})^T
\end{split}
\end{equation}

\balance
\bibliographystyle{IEEEtran}
\bibliography{library}

\begin{thebibliography}{10}
\providecommand{\url}[1]{#1}
\csname url@rmstyle\endcsname
\providecommand{\newblock}{\relax}
\providecommand{\bibinfo}[2]{#2}
\providecommand\BIBentrySTDinterwordspacing{\spaceskip=0pt\relax}
\providecommand\BIBentryALTinterwordstretchfactor{4}
\providecommand\BIBentryALTinterwordspacing{\spaceskip=\fontdimen2\font plus
\BIBentryALTinterwordstretchfactor\fontdimen3\font minus
  \fontdimen4\font\relax}
\providecommand\BIBforeignlanguage[2]{{%
\expandafter\ifx\csname l@#1\endcsname\relax
\typeout{** WARNING: IEEEtran.bst: No hyphenation pattern has been}%
\typeout{** loaded for the language `#1'. Using the pattern for}%
\typeout{** the default language instead.}%
\else
\language=\csname l@#1\endcsname
\fi
#2}}

\bibitem{Bircher2018}
A.~Bircher, M.~Kamel, K.~Alexis, H.~Oleynikova, and R.~Siegwart, ``{Receding
  horizon path planning for 3D exploration and surface inspection},''
  \emph{Autonomous Robots}, vol.~42, no.~2, pp. 291--306, 2018.

\bibitem{Dai_ICRA2020}
A.~Dai, S.~Papatheodorou, N.~Funk, D.~Tzoumanikas, and S.~Leutenegger, ``Fast
  frontier-based information-driven autonomous exploration with an {MAV},'' in
  \emph{IEEE Intl. Conf. on Robotics and Automation (ICRA)}, Paris, France,
  June 2020.

\bibitem{oleynikova2017voxblox}
H.~Oleynikova, Z.~Taylor, M.~Fehr, R.~Siegwart, and J.~Nieto, ``Voxblox:
  Incremental {3D} {Euclidean} signed distance fields for on-board {MAV}
  planning,'' in \emph{IEEE/RSJ Intl. Conf. on Intelligent Robots and Systems
  (IROS)}, 2017.

\bibitem{Hollinger2013}
G.~A. Hollinger, B.~Englot, F.~S. Hover, U.~Mitra, and G.~S. Sukhatme,
  ``{Active planning for underwater inspection and the benefit of
  adaptivity},'' \emph{Intl. J. of Robotics Research}, vol.~32, no.~1, pp.
  3--18, 2013.

\bibitem{Franz2016}
F.~S. Hover, R.~M. Eustice, A.~Kim, B.~Englot, H.~Johannsson, M.~Kaess, and
  J.~J. Leonard, ``Advanced perception, navigation and planning for autonomous
  in-water ship hull inspection,'' \emph{Intl. J. of Robotics Research},
  vol.~31, pp. 1445--1464, 2016.

\bibitem{bouman2020autonomous}
A.~Bouman, M.~F. Ginting, N.~Alatur, M.~Palieri, D.~D. Fan, T.~Touma,
  T.~Pailevanian, S.-K. Kim, K.~Otsu, J.~Burdick, \emph{et~al.}, ``Autonomous
  {Spot}: Long-range autonomous exploration of extreme environments with legged
  locomotion,'' \emph{arXiv preprint arXiv:2010.09259}, 2020.

\bibitem{ebadi2020lamp}
K.~Ebadi, Y.~Chang, M.~Palieri, A.~Stephens, A.~Hatteland, E.~Heiden,
  A.~Thakur, N.~Funabiki, B.~Morrell, S.~Wood, \emph{et~al.}, ``Lamp:
  Large-scale autonomous mapping and positioning for exploration of
  perceptually-degraded subterranean environments,'' in \emph{2020 IEEE
  International Conference on Robotics and Automation (ICRA)}.\hskip 1em plus
  0.5em minus 0.4em\relax IEEE, 2020, pp. 80--86.

\bibitem{mangelson2020characterizing}
J.~G. Mangelson, M.~Ghaffari, R.~Vasudevan, and R.~M. Eustice, ``Characterizing
  the uncertainty of jointly distributed poses in the {Lie} algebra,''
  \emph{IEEE Transactions on Robotics}, vol.~36, no.~5, pp. 1371--1388, 2020.

\bibitem{dellaert2017gtsam}
F.~Dellaert and C.~Beall, ``{GTSAM 4.0},'' \emph{URL: https://bitbucket.
  org/gtborg/gtsam}, 2017.

\bibitem{bosse2003atlas}
M.~Bosse, P.~Newman, J.~Leonard, M.~Soika, W.~Feiten, and S.~Teller, ``An
  {Atlas} framework for scalable mapping,'' in \emph{IEEE Intl. Conf. on
  Robotics and Automation (ICRA)}, vol.~2.\hskip 1em plus 0.5em minus
  0.4em\relax IEEE, 2003, pp. 1899--1906.

\bibitem{Nieto2006DenseSLAM}
J.~Nieto, J.~Guivant, and E.~Nebot, ``Denseslam: Simultaneous localization and
  dense mapping,'' \emph{Intl. J. of Robotics Research}, vol.~25, no.~8, pp.
  711--744, 2006.

\bibitem{hornung2013octomap}
A.~Hornung, K.~M. Wurm, M.~Bennewitz, C.~Stachniss, and W.~Burgard, ``Octomap:
  An efficient probabilistic 3d mapping framework based on octrees,''
  \emph{Autonomous Robots}, vol.~34, no.~3, pp. 189--206, 2013.

\bibitem{ho2018virtual}
B.-J. Ho, P.~Sodhi, P.~Teixeira, M.~Hsiao, T.~Kusnur, and M.~Kaess, ``Virtual
  occupancy grid map for submap-based pose graph slam and planning in 3d
  environments,'' in \emph{IEEE/RSJ Intl. Conf. on Intelligent Robots and
  Systems (IROS)}.\hskip 1em plus 0.5em minus 0.4em\relax IEEE, 2018, pp.
  2175--2182.

\bibitem{Sodhi2019}
P.~Sodhi, B.-J. Ho, and M.~Kaess, ``Online and consistent occupancy grid
  mapping for planning in unknown environments,'' in \emph{IEEE/RSJ Intl. Conf.
  on Intelligent Robots and Systems (IROS)}, 2019, pp. 7879--7886.

\bibitem{reijgwart2020voxgraph}
V.~{Reijgwart}, A.~{Millane}, H.~{Oleynikova}, R.~{Siegwart}, C.~{Cadena}, and
  J.~{Nieto}, ``Voxgraph: Globally consistent, volumetric mapping using signed
  distance function submaps,'' \emph{{IEEE} Robotics and Automation Letters},
  2020.

\bibitem{wang2020elastic}
Y.~Wang, N.~Funk, M.~Ramezani, S.~Papatheodorou, M.~Popovic, M.~Camurri,
  S.~Leutenegger, and M.~Fallon, ``Elastic and efficient lidar reconstruction
  for large-scale exploration tasks,'' \emph{IEEE International Conference on
  Robotics and Automation (ICRA)}, 2021.

\bibitem{Bellicoso2018AdvancesIR}
D.~Bellicoso, M.~Bjelonic, L.~Wellhausen, K.~Holtmann, F.~Guenther,
  M.~Tranzatto, P.~Fankhauser, and M.~Hutter, ``Advances in real-world
  applications for legged robots,'' \emph{J. of Field Robotics}, vol.~35, pp.
  1311--1326, 2018.

\bibitem{turner2014floor}
E.~Turner and A.~Zakhor, ``Floor plan generation and room labeling of indoor
  environments from laser range data,'' in \emph{{IEEE} Intl. Conf. on Computer
  Graphics Theory and Applications (GRAPP)}, 2014, pp. 1--12.

\bibitem{armeni20163d}
I.~Armeni, O.~Sener, A.~R. Zamir, H.~Jiang, I.~Brilakis, M.~Fischer, and
  S.~Savarese, ``3d semantic parsing of large-scale indoor spaces,'' in
  \emph{Proc. {IEEE} Int. Conf. Computer Vision and Pattern Recognition}, 2016,
  pp. 1534--1543.

\bibitem{mura2014automatic}
C.~Mura, O.~Mattausch, A.~J. Villanueva, E.~Gobbetti, and R.~Pajarola,
  ``Automatic room detection and reconstruction in cluttered indoor
  environments with complex room layouts,'' \emph{Computers \& Graphics},
  vol.~44, pp. 20--32, 2014.

\bibitem{mura2016piecewise}
C.~Mura, O.~Mattausch, and R.~Pajarola, ``Piecewise-planar reconstruction of
  multi-room interiors with arbitrary wall arrangements,'' in \emph{Computer
  Graphics Forum}, vol.~35, no.~7.\hskip 1em plus 0.5em minus 0.4em\relax Wiley
  Online Library, 2016, pp. 179--188.

\bibitem{ochmann2019automatic}
S.~Ochmann, R.~Vock, and R.~Klein, ``Automatic reconstruction of fully
  volumetric {3D} building models from oriented point clouds,'' \emph{ISPRS J.
  of Photogrammetry and Remote Sensing}, vol. 151, pp. 251--262, 2019.

\bibitem{Nikoohemat2020Routing}
S.~Nikoohemat, A.~A. Diakité, S.~Zlatanova, and G.~Vosselman, ``Indoor {3D}
  reconstruction from point clouds for optimal routing in complex buildings to
  support disaster management,'' \emph{Automation in Construction}, vol. 113,
  p. 103109, 2020.

\bibitem{Ramezani2020LiDARSLAM}
M.~{Ramezani}, G.~{Tinchev}, E.~{Iuganov}, and M.~{Fallon}, ``Online
  {LiDAR-SLAM} for legged robots with robust registration and deep-learned loop
  closure,'' in \emph{IEEE Intl. Conf. on Robotics and Automation (ICRA)},
  2020, pp. 4158--4164.

\bibitem{Pomerleau12comp}
F.~Pomerleau, F.~Colas, R.~Siegwart, and S.~Magnenat, ``{Comparing ICP Variants
  on Real-World Data Sets},'' \emph{Autonomous Robots}, vol.~34, no.~3, pp.
  133--148, Feb. 2013.

\bibitem{Simona2018Overlap}
S.~Nobili, G.~Tinchev, and M.~Fallon, ``Predicting alignment risk to prevent
  localization failure,'' in \emph{IEEE Intl. Conf. on Robotics and Automation
  (ICRA)}, 2018, pp. 1003--1010.

\bibitem{ramezani2020newer}
M.~Ramezani, Y.~Wang, M.~Camurri, D.~Wisth, M.~Mattamala, and M.~Fallon, ``{The
  Newer College Dataset}: Handheld {LiDAR}, inertial and vision with ground
  truth,'' in \emph{IEEE/RSJ Intl. Conf. on Intelligent Robots and Systems
  (IROS)}, 2020.

\bibitem{Lorensen1987MarchingCubes}
W.~E. Lorensen and H.~E. Cline, ``{Marching Cubes}: A high resolution {3D}
  surface construction algorithm,'' \emph{SIGGRAPH Comput. Graph.}, vol.~21,
  no.~4, p. 163–169, Aug. 1987.

\bibitem{klarsfeld1989baker}
S.~Klarsfeld and J.~Oteo, ``The {Baker-Campbell-Hausdorff} formula and the
  convergence of the {Magnus} expansion,'' \emph{Journal of physics A:
  mathematical and general}, vol.~22, no.~21, p. 4565, 1989.

\end{thebibliography}

\end{document}